\documentclass[letterpaper, 10 pt, conference]{ieeeconf}  

\IEEEoverridecommandlockouts                              

\usepackage{times}
\usepackage{epsfig}
\usepackage{graphicx}
\usepackage{amsmath}
\usepackage{amssymb}

\usepackage{multirow}
\usepackage{bm}
\usepackage{threeparttable}
\usepackage{algorithm}
\usepackage{algpseudocode}

\usepackage[breaklinks=true,bookmarks=false]{hyperref}
\hypersetup
{
colorlinks = true,
linkcolor = red,
anchorcolor = black,
citecolor = green,
urlcolor = magenta
}

\title{\LARGE \bf VLA-RAIL: A Real-Time Asynchronous Inference Linker \\ for VLA Models and Robots}
\author{Yongsheng Zhao$^{1}$, Lei Zhao$^{1}$, Baoping Cheng$^{1,*}$, Gongxin Yao$^{1}$, Xuanzhang Wen$^{1}$, Han Gao$^{1}$
\thanks{$^{1}$These authors are affiliated with China Mobile (Hangzhou) Information Technology Co., Ltd., Hangzhou, 310023, China. (Baoping Cheng is the corresponding author, email: chengbaoping@cmhi.chinamobile.com)}%
}

\begin{document}

\maketitle
\thispagestyle{empty}
\pagestyle{empty}

\begin{abstract}
Vision-Language-Action (VLA) models have achieved remarkable breakthroughs in robotics, with the action chunk playing a dominant role in these advances. Given the real-time and continuous nature of robotic motion control, the strategies for fusing a queue of successive action chunks have a profound impact on the overall performance of VLA models. Existing methods suffer from jitter, stalling, or even pauses in robotic action execution, which not only limits the achievable execution speed but also reduces the overall success rate of task completion. This paper introduces \textbf{VLA-RAIL} (A \textbf{R}eal-Time \textbf{A}synchronous \textbf{I}nference \textbf{L}inker), a novel framework designed to address these issues by conducting model inference and robot motion control asynchronously and guaranteeing smooth, continuous, and high-speed action execution. The core contributions of the paper are two fold: a \textbf{\textit{Trajectory Smoother}} that effectively filters out the noise and jitter in the trajectory of one action chunk using polynomial fitting and a \textbf{\textit{Chunk Fuser}} that seamlessly align the current executing trajectory and the newly arrived chunk, ensuring position, velocity, and acceleration continuity between two successive action chunks. We validate the effectiveness of VLA-RAIL on a benchmark of dynamic simulation tasks and several real-world manipulation tasks. Experimental results demonstrate that VLA-RAIL significantly reduces motion jitter, enhances execution speed, and improves task success rates, which will become a key infrastructure for the large-scale deployment of VLA models.
\end{abstract}

\section{Introduction}

The advent of large, pre-trained models \cite{zhu2024survey, zhang2024vision} has given rise to a new category of general-purpose robotic policies, referred to as Vision-Language-Action (VLA) models \cite{sapkota2025vision}. Built on large and diverse datasets, these models enable robots to understand natural language commands, extract spatial-visual semantics, and produce actions suited to open-world manipulation tasks. Despite the conceptual uniformity of the VLA paradigm, existing models vary widely in input modality, model architecture, output format, and inference process. At the same time, robotic platforms \cite{zhu2024overview, fan2024review, tong2024advancements} also exhibit substantial heterogeneity in both hardware composition and software interface. Therefore, deploying VLA models on physical robots remains closely coupled to particular robot types and hardware configurations, which substantially limits their scalability and cross-platform portability. 

\begin{figure}[t]
  \centering
  \includegraphics[width=1\linewidth]{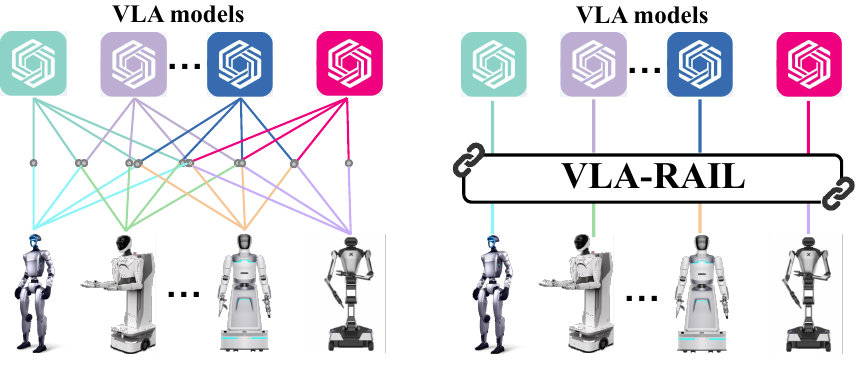}
  \caption{Illustration of current issue in VLA deployment (left) and the advantage of VLA-RAIL (right). Deploying VLA models on heterogeneous robots involves one-to-one adaptation, while VLA-RAIL provides a unified framework that enables plug-and-play deployment across diverse VLA models and heterogeneous robots.}
  \label{fig:abstract}
\end{figure}

Based on the generalized multi-modal representations of Vision-Language Models \cite{zhang2024vision, liu2025survey}, VLA models further integrate action experts \cite{sapkota2025vision} to achieve generalized robotic manipulation. To balance the inference latency of these large models with real-time motor control, the action chunking technique \cite{zhao2023learning} is applied to predict multiple frames of actions within a single inference cycle. Although this trajectory-level prediction has become the dominant paradigm for action experts in VLA models \cite{kim2024openvla, black2024pi_0, liurdt, bu2025agibot_iros, bjorck2025gr00t}, a brief pause still occurs at the end of each action chunk because action inference and execution alternate sequentially.

Beyond the inherent capabilities of VLA models, the real-world performance on physical robots is critically influenced by three external factors. 1) In real-world manipulation, proprioceptive and visual observations are updated concurrently during action execution, necessitating asynchronous and proactive model inference to avoid pause-and-go behavior. However, variable latency \cite{black2025real, ma2025running, sendai2025leave} introduced by cameras, networks, and GPUs makes precise temporal alignment across successive action chunks both critical and difficult. Such misalignment leads to discontinuities at action transitions, degrading control accuracy and motion stability. 2) VLA models are typically trained on action trajectories collected through human teleoperation \cite{darvish2023teleoperation}, which inevitably contain jitter and introduce noise into prediction. Moreover, a generative action model based on Flow Matching \cite{lipman2023flow} or Diffusion Policy \cite{chi2025diffusion} can even exacerbate this phenomenon. 3) Action chunks are independently learned by the VLA model from proprioceptive and visual observations at different time steps, without explicit trajectory constraints on position, velocity, or acceleration across successive chunks. Consequently, even with correct temporal alignment, trajectory discontinuities may still arise at chunk boundaries, resulting in non-smooth motion. Therefore, many VLA demonstrations slow down the execution of action chunks to reduce the impact of jitter, sometimes by a factor of 4 to 8.

To address the aforementioned problems, we propose VLA-RAIL (\textbf{R}eal-Time \textbf{A}synchronous \textbf{I}nference \textbf{L}inker), a general and scalable asynchronous inference framework that establishes a linker between multiple VLA models and heterogeneous robots, as shown in Fig. \ref{fig:abstract}. This plug-and-play framework decouples VLA models and robotic hardware using a client-server architecture. The client provides a unified interface that abstracts  heterogeneous robots, receiving proprioceptive and visual observations from the robot while dispatching motion control commands to the robot. The VLA models run on the server, communicating with the client via a request–response pattern over the ZMQ \cite{hintjens2013zeromq} protocol. 

In addition, VLA-RAIL incorporates a two-stage post-processing strategy for action chunks to effectively eliminate motion jitters. Specifically, the intra-chunk stage applies trajectory smoothing to effectively eliminate jitter within a chunk. The inter-chunk stage utilizes temporal alignment and a chunk fusion method to further eliminate the abrupt transitions between successive chunks. Moreover, VLA-RAIL also employs a real-time data manager to temporally align the proprioception state and visual observation, as well as a multi-thread architecture to handle perception, inference, trajectory post-processing and robot control concurrently. Based on this decoupled computation and control pipeline, task execution can be accelerated beyond teleoperation speed by jointly tuning the action trajectory interpolation and command dispatch frequencies. Extensive real-world experiments demonstrate that VLA-RAIL can produce smoother action trajectories, enabling VLA models to perform more precise and efficient robotic manipulation.

In conclusion, our main contributions are:

\begin{itemize}
    \item An open-source and model-agnostic asynchronous framework for VLA model inference, serving as a plug-and-play middleware to seamlessly link diverse VLA models and heterogeneous robotic platforms.
    \item A two-stage action chunk post-processing strategy consists of intra-chunk smoothing and inter-chunk fusion, which eliminates the motion jitter arising from prediction noise and asynchronous temporal misalignment, thereby improving the overall task success rate.
    \item A simple yet efficient action acceleration strategy that can speed up execution to reach the hardware limits by jointly tuning the action trajectory interpolation and command dispatch frequencies. 
\end{itemize}

\section{Related Work}
\subsection{Vision-Language-Action Models}
Vision-Language-Action (VLA) models have emerged as a promising paradigm \cite{sapkota2025vision} to unify visual perception, natural language understanding, and robot control within a single policy architecture for embodied AI \cite{liu2025aligning}. Early works attempted to train action policies from scratch using real robot trajectories via imitation learning, with RVT \cite{goyal2023rvt} adopting a standard vision transformer and Diffusion Policy \cite{chi2023diffusion} adopting a conditional denoising diffusion transformer. However, since the scarcity of robot demonstration data limits the generalization of these approaches, recent VLA models increasingly build on pretrained vision–language foundations \cite{zhang2024vision} rather than training policies from scratch. Examples such as OpenVLA \cite{kim2024openvla}, RDT \cite{liurdt}, GR00T \cite{bjorck2025gr00t}, Pi \cite{black2024pi_0}, and GO1 \cite{bu2025agibot_iros} embed large vision–language models directly into their architectures, allowing them to transfer rich knowledge from internet-scale data and finetune on limited amounts of real robot demonstration data. Furthermore, action chunking \cite{zhao2023learning} has become the dominant paradigm for VLA models, which predicts multi-frame actions in a single inference. It allows the models to capture trajectory-level motion intentions, alleviating the challenge of learning high-frequency and low-level motor control. Nevertheless, deploying existing models on heterogeneous robots often requires substantial adaptation, highlighting the urgent need for a general tool to improve cross-platform portability.

\subsection{Action Chunk Fusion Methods}
The most straightforward approach is to run VLA models in a synchronous loop \cite{sendai2025leave}: the robot first acquires observations, infers action chunk, executes actions, and only then starts the next inference cycle. In this case, successive chunks can transition naturally, yet the necessary waiting time interval results in halting of motion. Although asynchronous inference \cite{sendai2025leave} eliminates idle waiting time through parallel execution and inference, it requires additional strategies to fuse temporally overlapping regions between successive action chunks. Specifically, some method \cite{shukor2025smolvla} simply switches to the next action chunk, overlooking the temporal misalignment that induces sharp discontinuities between chunks and leads to unstable robot motion. Real-time Chunking (RTC) \cite{black2025real} fuses successive action chunks by freezing the first few actions in the new chunk and inpainting the remaining actions in the current chunk. A customized soft mask strategy applies exponentially decaying guidance weights to the overlapping region, but it complicates deployment and lacks transferability across different VLA models. A2C2 \cite{sendai2025leave} learns a dynamic compensation network to rectify asynchronous action errors, but it requires architectural modifications and training of the additional branch. VLASH \cite{tang2025vlash} estimates the future robot proprioception state during inference and conditions the VLA models on it to predict future-aware actions for smooth chunk fusion. However, the requirement for dataset reorganization and model retraining significantly hinders plug-and-play deployment.

\section{Preliminaries and Problem Formulation}
\label{sec:problem}

In this section, we formally define the action-chunk based VLA policy and model the temporal correlation among robot observation acquisition, VLA model inference, and motion control within a real-time pipeline . Finally, we provide a mathematical characterization of intra-chunk trajectory jitter and inter-chunk unconstrained discontinuities.

\begin{figure*}[htbp]
  \centering
  \includegraphics[width=\textwidth]{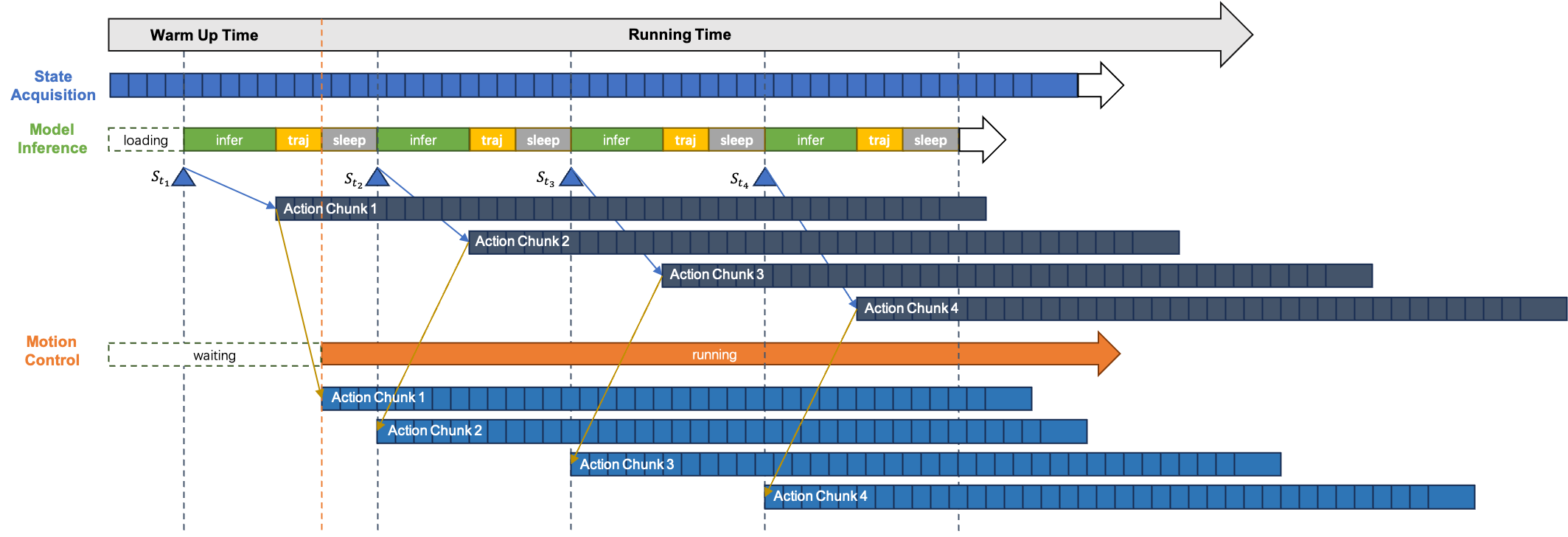}
  \caption{Temporal Pipeline Of Robotic Manipulation. There are three real-time pipelines happened concurrently during robotic manipulation execution. The first one, named eye pipeline, is the state retrieval pipeline that acquires the proprioception data and visual images via hardware drivers at a frequency of $f_{state}$. The second one, name brain pipeline, is the model inference pipeline that predicts new action chunks using current acquired state data at a frequency of $f_{infer}$. The third one, named hand pipeline, is the motion control pipeline that sends the action command to robot at a frequency of $f_{ctrl}$. }
  \label{fig:pipeline}
\end{figure*}

\subsection{Mathematical Formulation Of VLA Policy}
Generally, an action-chunk based VLA model takes the language instruction $l$, visual observation $o_t \in \mathcal{O}$ and proprioception state $q_t \in \mathcal{Q}$ as input and outputs a chunk of actions $\mathbf{A}_t^H$, where $t$ denotes the time, $\mathcal{O}$ denotes the RGB images, $\mathcal{Q}$ denotes the joint and end-effector states and $H$ denotes the prediction horizon (chunk size). Therefore, a VLA policy $\pi_\theta$ can be formulated as:
\begin{equation}
    \mathbf{A}_t^H = \pi_\theta(l, q_t, o_t) = [a_t^0, a_t^1, \dots, a_t^{H-1}]
\end{equation}
where $a_t^k \in \mathcal{A}$ represents the $k$-th predicted target action at time $t$. As we can see, two consecutive action chunks $\mathbf{A}_{t}^H$ and $\mathbf{A}_{t^{'}}^H$ are independently inferred from $(l, q_t, o_t)$ and $(l, q_{t^{'}}, o_{t^{'}})$ without explicit constraints enforcing trajectory continuity and consistency of velocity or acceleration.

\subsection{Temporal Pipeline Of Robotic Manipulation}
As depicted in \mbox{Fig. \ref{fig:pipeline}}, generalized robotic manipulation is a complex \textbf{hand–eye–brain} coordination process, involving three real-time pipelines that operate simultaneously. Given the robot’s real-time state data $(q_t, o_t)$ at time $t$ acquired by the eye pipeline, the $\mathbf{A}_t^H$ will be predicted by brain pipeline at time $\hat{t}$.
\begin{equation}
    \hat{t} = t + \Delta t_{i} + \Delta t_{t}
\end{equation}
where $\Delta t_{i}$ denotes the inference duration of VLA model, $\Delta t_{t}$ denotes the trajectory post-process duration. 

In a synchronous execution mode, the robot has to halt and wait for latency $\Delta t_{infer} + \Delta t_{traj}$ to get the next action chunk to execute. To address the pause issue, an asynchronous execution mode is preferred, where the next action chunk is predicted asynchronously while executing the current one. However, the predicted action chunk should be temporally aligned to the current one according to the time when the robot's state happens. The alignment time can be formulated as:
\begin{equation}
    \tilde{t_{a}} = \Delta t_{i} + \Delta t_{t} +  \Delta t_{o}
    \label{eq:t_offset}
\end{equation}
Where $\Delta t_{o}$ denotes the time the sensors used to process signals and return observations. Generally, it can't be estimated accurately for most robotic platforms.

The hand pipeline sends actions of the post-processed trajectories frame-by-frame to the robot at a frequency of $f_{ctrl}$ for execution. Note that the policy is typically queried at a frequency $f_{infer}$ significantly lower than the robot's low-level control frequency $f_{ctrl}$.


\subsection{The Analysis Of Trajectory Jitter}
There are two kinds of trajectory jitter in action-chunk-based robotic manipulation  as follows:

\textbf{Inter-Chunk Discontinuity:} The trajectory discontinuity in the transition between two consecutive action chunks is caused by the isolated action chunk modeling and temporal misalignment, as illustrated by the preceding subsections. Suppose that there are two action chunks $\mathbf{A}_{t_0}^H$ and $\mathbf{A}_{t_1}^H$, where $t_0$ and $t_1$ are the time when the robot's state is acquired. Then we can quantize the discontinuity using the following equation:
\begin{eqnarray}
    \Delta \mathbf{A}^{t_{s}} &=& \mathbf{A}_{t_0}^{t_{s}} - \mathbf{A}_{t_1}^{t_{a}}  \\ 
    &=& \pi_\theta(l, q_{t_0}, o_{t_0})^{t_{s}} - \pi_\theta(l, q_{t_1}, o_{t_1})^{t_{a}} \nonumber
\end{eqnarray}
where $t_{s}$ denotes the time we switch to $\mathbf{A}_{t_1}^H$ from $\mathbf{A}_{t_0}^H$, and $t_1-t_0+t_{a} \le t_{s}$.

\begin{figure*}[htbp]
  \centering
  \includegraphics[width=\textwidth]{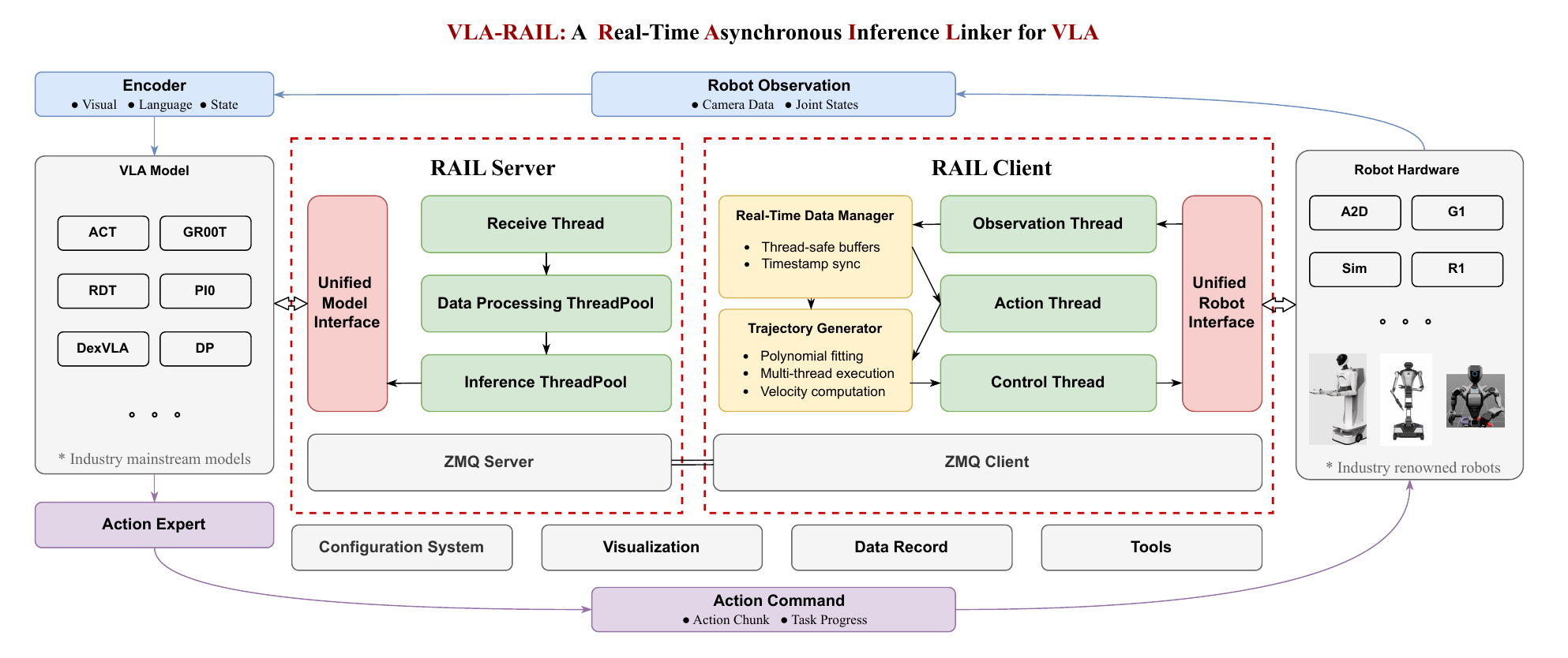}
  \caption{Overview of VLA-RAIL. The asynchronous framework decouples VLA model inference with robotic motion control using a client-server architecture, with request–response communication implemented via ZMQ protocol. By exposing unified interfaces on both the server and client sides, VLA-RAIL supports plug-and-play deployment across multiple VLA models and heterogeneous robotic platforms. Functionally, the server hosts computationally intensive VLA model inference on GPU devices, while the client employs a multi-threaded design to concurrently handle proprioceptive and visual observation acquisition, model inference requests, trajectory post-processing, and real-time robot motion control. }
  \label{fig:architecture}
\end{figure*}

\textbf{Intra-Chunk Noise:} Trajectory noise arises from human teleoperation and is introduced into the supervision used for training the model. Usually, the loss function of a VLA model is based on the Manhattan distance ($\mathcal{L}_1$) or the Euclidean distance ($\mathcal{L}_2$).
\begin{eqnarray}
    \mathcal{L}_1 = \frac{1}{H}\sum_{i=1}^{H} \left| \tilde{x}_i - x_i \right| \\
    \mathcal{L}_2 = \frac{1}{H}\sum_{i=1}^{H} \left| \tilde{x}_i - x_i \right|^2
\end{eqnarray}
Where $H$ denotes the size of action chunk, $\tilde{x}_i$ denotes the $i-th$ predicted value of one joint and $x_i$ denotes the $i-th$ ground truth value.
Let $\Delta x_i$ be a real positive value denoting the $i-th$ predicted error of one joint, the $\mathcal{L}_1$ or $\mathcal{L}_2$ loss will be the same in the following scenario:
\begin{equation}
\begin{cases}
\tilde{x}_i = x_i + \Delta x_i \\
\tilde{x}_{i+1} = x_{i+1} - \Delta x_{i+1}

\end{cases}
\end{equation}
However, the trajectory jitter is amplified by $\Delta x_i + \Delta x_{i+1}$. As we can see, the loss functions used in the training scheme of VLA model do not effectively address the issue of trajectory jitter.

\section{The VLA-RAIL Framework}

\subsection{Overview}
We propose VLA-RAIL, a general framework addressing the issues formulated in Section \ref{sec:problem}. As illustrated in \mbox{Fig. \ref{fig:architecture}}, it adopts a client-server architecture that decouples VLA model inference from robotic motion control. VLA-RAIL enables parallel action computation and robot control via sophisticated data and process orchestration, thereby eliminating idle latency during VLA model inference. Specifically, the client maintains three concurrent processes for real-time robotic manipulation: 1) an \textit{eye process} managing real-time  proprioceptive and visual observations from robot at frequency $f_{obs}$; 2) a \textit{brain process} asynchronously communicating with the server to query and post-process new action chunk; and 3) a \textit{hand process} dispatching commands at frequency $f_{ctrl}$. To eliminate the motion jitter arising from prediction noise and asynchronous temporal misalignment, we introduce a trajectory post-processing pipeline as the core algorithmic contribution. This pipeline consists of two stages: \textit{intra-chunk smoothing}, which filters trajectory noise within each chunk, and \textit{inter-chunk fusion}, which ensures seamless transitions between successive chunks.

\subsection{Intra-Chunk Trajectory Smoothing}

Action chunks predicted by VLA often exhibit high-frequency oscillations due to teleoperation noise in training data, stochastic sampling in generative models, and discretization artifacts. We effectively address this issue using a time-parameterized continuous model to represent the trajectories in action chunk, such as multi-order polynomials or Fourier series. Constrained by maximum velocity and acceleration limits, the trajectories could be well represented by three- or higher-order polynomials.

For each joint $j$, let $\{(t_k, a_t^k)\}_{k=0}^{H-1}$ denote the $H$ trajectory waypoints. We parameterize the action trajectory using a $d$-th order polynomial as:
\begin{equation}
    \mathcal{P}_j(t; \mathbf{c}) = \sum_{i=0}^{d} c_i t^i
\end{equation}
The optimal coefficients are identified via least squares estimation to minimize the Euclidean distance between the waypoints and polynomial. Given the Vandermonde matrix $\mathbf{V} \in \mathbb{R}^{H \times (d+1)}$ with entries $V_{ki} = t_k^i$ and the waypoints $\mathbf{y} = [a_t^0, \ldots, a_t^{H-1}]^\top$, the closed-form solution is:
\begin{equation}
    \mathbf{c}^* = (\mathbf{V}^\top \mathbf{V})^{-1} \mathbf{V}^\top \mathbf{y}
    \label{eq:least_squares}
\end{equation}

The order $d$ of the polynomial controls the smoothness–fidelity trade-off: higher degrees preserve more detail but retain noise, while lower degrees act as low-pass filters. Taking into account both computational efficiency and trajectory smoothness, we employ cubic polynomials \mbox{($d=3$)}. The matrix $\mathbf{V}^\top \mathbf{V} \in \mathbb{R}^{4 \times 4}$ can be solved in sub-millisecond latency, allowing real-time response.

\subsection{Inter-Chunk Seamless Fusion}

As illustrated in Section \ref{sec:problem}, the key to seamless inter-chunk fusion lies in two aspects: precise temporal alignment and trajectory smoothing optimization.

\textbf{Temporal Alignment:} The two action chunks must be aligned based on the timing of event occurrences, as computed in Equation \ref{eq:t_offset}. Since $\Delta t_{o}$ cannot be accurately estimated, we reformulate this problem as an optimization task subject to motion consistency constraints.
\begin{equation}
    \begin{aligned}
    \max_{\boldsymbol{t_{a}}}\sum_{i=0}^{m}sign(({A}_{t_{1}}^{t_{a}}[i]-{A}_{t_{0}}^{t_{s}}[i]) \frac{\partial{A}_{t_{0}}^{t_{s}}[i]}{\partial t}) \\
    \text{s.t.} \quad
    t_{a} \in (0,t_{w}),\quad sign(x) = \begin{cases}
    1 & x \geq 0 \\
    -1 & x < 0
    \end{cases}
    \end{aligned}
    \label{eq:temporal_alignment}
\end{equation}
where $t_w$ denotes the time window for temporal alignment and $m$ denotes the dimension of action.

\textbf{Trajectory Smoothing:} Quintic spline interpolation is commonly used for smooth trajectory generation under position, velocity, and acceleration constraints at the start and end points, which guaranties minimum jerk over the whole trajectory smoothness. However, it has overshoot problem due to Runge's Phenomenon of high-order polynomials, which results in shaking motion of robot execution. We propose a dual-quintic spline interpolation method to effectively address this issue. Let $\mathcal{Q}_l^{t}[i] = \sum_{j=0}^{5} b^l_j \tau^j$ and $\mathcal{Q}_r^{t}[i]  = \sum_{j=0}^{5} b^r_j \tau^j$ denotes two quintic splines,  the dual-quintic spline interpolation can be formulated as follows.
\begin{align}
    &\begin{cases}
    \mathcal{Q}_l^0[i] = {A}_{t_{0}}^{t_{s}}[i], & \mathcal{Q}_l^{\frac{t_{q}}{2}}[i] = \frac{{A}_{t_{0}}^{t_{s}}[i]+{A}_{t_{1}}^{t_{a}+t_{q}}[i]}{2} \\
    \dot{\mathcal{Q}_l}^0[i] = \dot{{A}}_{t_{0}}^{t_{s}}[i], & \dot{\mathcal{Q_l}}^{\frac{t_{q}}{2}}[i] = \frac{\dot{{A}}_{t_{0}}^{t_{s}}[i]+\dot{{A}}_{t_{1}}^{t_{a}+t_{q}}[i]}{2} \\
    \ddot{\mathcal{Q}_l}^0[i] = \ddot{{A}}_{t_{0}}^{t_{s}}[i], & \ddot{\mathcal{Q}_l}^{\frac{t_{q}}{2}}[i] = 0
    \end{cases} \\
    &\begin{cases}
    \mathcal{Q}_r^0[i] = \frac{{A}_{t_{0}}^{t_{s}}[i]+{A}_{t_{1}}^{t_{a}+t_{q}}[i]}{2}, & \mathcal{Q}_r^{\frac{t_{q}}{2}}[i] = {A}_{t_{1}}^{t_{a}+t_{q}}[i] \\
    \dot{\mathcal{Q}_r}^0[i] = \frac{\dot{{A}}_{t_{0}}^{t_{s}}[i]+\dot{{A}}_{t_{1}}^{t_{a}+t_{q}}[i]}{2}, & \dot{\mathcal{Q}_r}^{\frac{t_{q}}{2}}[i] = \dot{{A}}_{t_{1}}^{t_{a}+t_{q}}[i] \\
    \ddot{\mathcal{Q}_r}^0[i] = 0, & \ddot{\mathcal{Q}_r}^{\frac{t_{q}}{2}}[i] = \ddot{{A}}_{t_{1}}^{t_{a}+t_{q}}[i] 
    \end{cases}
\end{align}

Where $t_q$ denotes the time window for interpolation. The coefficients of $\mathcal{Q}_l^{t}[i]$ and $\mathcal{Q}_r^{t}[i]$ are obtained by solving a $6 \times 6$ linear system.

The composite trajectory is thus:
\begin{equation}
    \mathcal{A}^t[i] = \begin{cases}
        {A}_{t_{0}}^{t}[i] & t < t_{s} \\
        \mathcal{Q}_l^{t-t_{s}}[i] & t_{s} \leq t < t_{s} + \frac{t_{q}}{2} \\
        \mathcal{Q}_r^{t-t_{s}-\frac{t_{q}}{2}}[i] & t_{s} +\frac{t_{q}}{2} \leq t < t_{s} + t_{q} \\
        {A}_{t_{1}}^{t-t_s+t_a}[i] & t \geq t_{s} + t_{q}
    \end{cases}
\end{equation}

\subsection{Execution Acceleration}

One advantage of VLA-RAIL is the ability to accelerate robotic execution speed without retraining the VLA model. Usually, the frequency $f_{act}$ of action chunk predicted by VLA model is the same as the frequency $f_{obs}$ (typically $30Hz$) of demonstration dataset collected via teleoperation. The frequency $f_{act}$ is far lower than the required frequency $f_{ctrl}$ (typically $\ge 100Hz$) of robotic motion control. Therefore, waypoints must be interpolated at a higher frequency $f_{interp}$.

We can finely control the robot’s execution speed by adjusting the frequency of $f_{interp}$ and $f_{ctrl}$. The acceleration ratio $\alpha$ is defined by:
\begin{equation}
    \alpha = \frac{f_{ctrl}}{f_{interp}}
\end{equation}

If $\alpha > 1$, the action execution speed is faster than the teleoperation speed. The interpolation frequency $f_{interp}$ can be any value because the policy after inter-chunk fusion is a continuous model of variable $t$. The control frequency $f_{ctrl}$ is limited by hardware capabilities, such as communication bandwidth and low level motor control policy.

\subsection{Algorithm Summary}

Algorithm \ref{alg:vla_rail} presents the complete VLA-RAIL control pipeline. The algorithm operates as a real-time control loop running at frequency $f_{ctrl}$. At each iteration, it checks for newly arrived action chunks from the inference server. When a new chunk arrives, the algorithm first performs temporal alignment by computing the latency-compensated index $k^*$, discarding stale actions that correspond to past robot states. It then applies intra-chunk smoothing via cubic polynomial fitting. If a previous trajectory exists, a quintic blending polynomial is constructed to ensure $\mathcal{C}^2$ continuity during the transition; otherwise, the new trajectory is directly activated. Finally, the active trajectory is evaluated at the current time and the resulting command is dispatched to the robot.

\begin{algorithm}[t]
\caption{VLA-RAIL Trajectory Control}
\label{alg:vla_rail}
\begin{algorithmic}[1]
\Require Control frequency $f_{ctrl}$, blend duration $T_{blend}$
\State Initialize $\mathcal{T}_{active} \gets$ null
\While{system active}
    \State $t \gets$ GetCurrentTime()
    \If{NewChunkAvailable($\mathbf{A}_{new}$, $t_{obs}$)}
        \State $k^* \gets \lfloor (t - t_{obs}) \cdot f_{ctrl} \rfloor$ \Comment{Temporal alignment}
        \State $\mathcal{P}_{new} \gets$ FitCubic($\mathbf{A}_{new}[k^*:]$) \Comment{Intra-chunk smoothing}
        \If{$\mathcal{T}_{active} \neq$ null}
            \State $\mathcal{Q} \gets$ SolveQuintic($\mathcal{T}_{active}$, $\mathcal{P}_{new}$, $t$, $T_{blend}$)
            \State $\mathcal{T}_{active} \gets$ Compose($\mathcal{Q}$, $\mathcal{P}_{new}$)
        \Else
            \State $\mathcal{T}_{active} \gets \mathcal{P}_{new}$
        \EndIf
    \EndIf
    \State SendToRobot($\mathcal{T}_{active}(t)$)
\EndWhile
\end{algorithmic}
\end{algorithm}

\section{Experiments}

We have conducted extensive experiments on real-world manipulation tasks to systematically validate four key advantages of VLA-RAIL:
1) improved trajectory smoothness over existing methods;
2) increased task success rates resulting from smoother trajectories;
3) broad compatibility across diverse VLA models; and
4) accelerated execution achieved without policy retraining.

\begin{figure*}[htbp]
  \centering
  \includegraphics[width=1\linewidth]{images/exp_traj.pdf}
  \caption{Trajectory smoothness comparison under three inference-execution strategies: without post-processing (Trajectory-1), naive switching \mbox{(Trajectory-2)}, and VLA-RAIL (Trajectory-3). Left column: time-series plots of joint angle (top), velocity (middle), and acceleration (bottom). The zoomed insets highlight trajectory details, with the right inset directly comparing the original VLA output and VLA-RAIL post-processed result. Right column: standard deviation (std) of each metric computed at 1-second intervals, where lower values indicate smoother motion. }
  \label{fig:exp_traj}
\end{figure*}

\subsection{Experimental Setup}


\textbf{Hardware.} The experiments are conducted on an AgiBot G1 robot \cite{ZhiyuanG1}, which consists of dual arms with 14-DoF joints and 2 parallel-jaw grippers. Both the VLA-RAIL server and client are deployed on a computer equipped with NVIDIA RTX 4080 GPU (Laptop version, 12 GB VRAM). This computer is connected to the real robot, ensuring that all computation and execution sides are on the same local area network to minimize communication latency.

\textbf{VLA Models.} VLA-RAIL is model-agnostic and currently compatible with the following VLA models: GO1 \cite{bu2025agibot_iros}, SmolVLA \cite{shukor2025smolvla}, $\pi_0$ \cite{black2024pi_0}, $\pi_{0.5}$ \cite{black2024pi_0}, GR00T \cite{bjorck2025gr00t}, and others. In our experiments, all models are trained on the same teleoperated demonstration dataset collected at 30Hz. Subsequently, they are employed to drive robot operations through our VLA-RAIL framework.

\textbf{Baselines.} To demonstrate the advantages of VLA-RAIL, we compare against two other inference–execution strategies: 1) direct execution of raw VLA model outputs without any additional post-processing (denoted as \textbf{w/o post-processing}), and 2) direct switching between consecutive action chunks with temporal alignment (denoted as \textbf{naive switching}). Unlike previous work \cite{shukor2025smolvla} similar to naive switching, polynomial fitting is applied within each action chunk. Notably, both baselines adopt an asynchronous inference–execution paradigm. It is widely acknowledged that synchronous strategy \cite{black2025real, tang2025vlash} often results in lower execution efficiency as well as robot idling and motion jitter. Therefore, it is not included in our experiments.

\textbf{Evaluation Metrics.} Performance is evaluated using the following metrics: 1) the task success rate over 20 real-robot trials; 2) the average completion time for successful trials, with complex tasks further decomposed by several stages; and 3) the trajectory smoothness metrics, including the standard deviation of joint angles, velocities, and accelerations computed at 1-second intervals.


\subsection{Trajectory Smoothness Analysis}

We first assess the kinematic quality of the action trajectories by analyzing the joint angle, velocity, and acceleration profiles. Fig. \ref{fig:exp_traj} shows three representative trajectories obtained by repeatedly performing the same robotic grasping task under three different inference–execution strategies.

\textbf{Joint Angle Trajectory:} The top-left panel illustrates the joint angle trajectories over time. Without post-processing, trajectory-1 exhibits noticeable high-frequency fluctuations throughout the motion, particularly evident in the zoomed-in view of the trajectory segment from 0 to 10 seconds. The naive switching strategy (Trajectory-2) reduces some fluctuations through intra-chunk polynomial fitting but still shows abrupt transitions at action chunk boundaries. In contrast, VLA-RAIL (Trajectory-3) produces a smooth, continuous curve that closely follows the intended motion profile while filtering out prediction noise. Specifically, the zoomed-in view of the trajectory segment from 14 to 20 seconds directly compares the original VLA model output (Trajectory-3 [Original]) with the VLA-RAIL post-processed result (Trajectory-3 [VLA-RAIL]), demonstrating the effectiveness of our smoothing algorithm in preserving the overall trajectory shape while eliminating prediction noise.

\textbf{Velocity Profile:} The middle-left panel shows the velocity profiles of the three strategies. The trajectory without post-processing demonstrates severe velocity fluctuations, with frequent sign reversals that indicate jerky, unstable motion. Naive switching strategy improves upon this but exhibits sharp velocity discontinuities at action chunk boundaries. VLA-RAIL maintains a smooth velocity profile with gradual transitions, achieving $\mathcal{C}^1$ continuity throughout the trajectory.

\textbf{Acceleration Analysis:} The bottom-left panel reveals the most striking differences among the three inference–execution strategies. The raw trajectory without post-processing exhibits sharp acceleration spikes, which may violate safety constraints and impose significant mechanical stress on the robot. The naive switching strategy significantly reduces the frequency of acceleration peaks. However, similar to the corresponding velocity profile, it still poses risks at the action chunk boundaries. In contrast, VLA-RAIL exhibits bounded and smooth accelerations, remaining within a narrow range and approaching $\mathcal{C}^2$ continuity, which is crucial for safe and efficient robot operation.

\textbf{Standard Deviation Analysis:} The right column of \mbox{Fig. \ref{fig:exp_traj}} displays the standard deviation (std) of joint angle, velocity, and acceleration computed at 1-second intervals. Across all three metrics, VLA-RAIL consistently achieves the lowest standard deviation values. In particular, the acceleration standard deviation for VLA-RAIL remains near zero throughout the entire trajectory, whereas the other strategies exhibit significantly higher fluctuations. This quantitative analysis confirms that VLA-RAIL effectively eliminates both intra-chunk noise and inter-chunk discontinuities, resulting in smoother real-robot motions.

\begin{figure}[t]
  \centering  
  \includegraphics[width=1\linewidth]{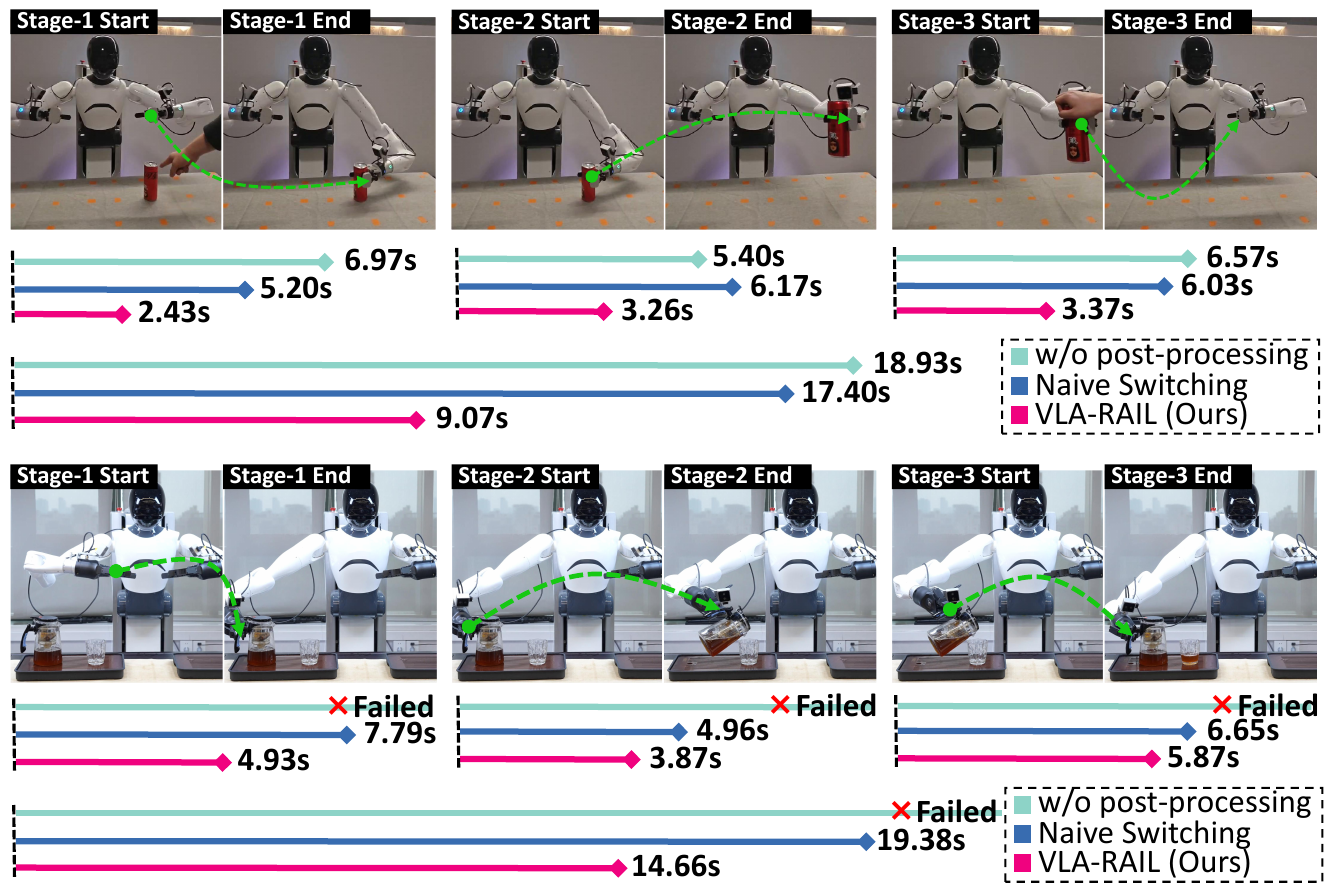}
  \caption{Illustrations of the task completion speed under different inference–execution strategies. The \textbf{bottle-grasping} task (top) includes 3 stage: 1) a human points to a bottle while the robot recognizes and grasps it; 2) the robot delivers the bottle to the human; and 3) the human takes the bottle and the robot returns to its initial pose. The \textbf{tea-pouring} task (bottom) includes 3 stages: 1) recognizing and grasping the teapot handle; 2) adjusting the teapot pose above the cup; and 3) pouring tea and returning the teapot to its original position. The horizontal bars shows the average time spent in each stage and the total task duration. Here, VLA-RAIL achieves up to $2.09\times$ speedup compared to the strategy without any post-processing.}
  \label{fig:exp_speed}
\end{figure}

\subsection{Task Completion Time}

VLA-RAIL can accelerate the execution of action chunks by jointly adjusting the trajectory interpolation and motor control frequencies. This allows VLA-driven robots to operate faster than human teleoperation, without the need to recollect high-frequency data and retrain the models. \mbox{Fig. \ref{fig:exp_speed}} shows two representative manipulation tasks, with each task divided into three consecutive stages, allowing a comparison of the completion times for different strategies.

\textbf{Bottle-Grasping Task:} The top portion of \mbox{Fig. \ref{fig:exp_speed}} illustrates a human-robot handover scenario. In Stage-1 (recognition and grasping), VLA-RAIL completes in 2.43s compared to 5.20s for Naive Switching and 6.97s for without post-processing, representing a 53\% and 65\% reduction respectively. Stage-2 (delivering to human) shows similar trends with VLA-RAIL requiring only 3.26s versus 6.17s (Naive Switching) and 5.40s (without post-processing). In Stage-3 (returning to initial pose), VLA-RAIL maintains its advantage with 3.37s compared to 6.03s and 6.57s for the other methods. The total task completion time demonstrates that VLA-RAIL finishes in 9.07s, compared to 17.40s for Naive Switching and 18.93s for without post-processing, achieving approximately $2.09\times$ speedup.

\textbf{Tea-Pouring Task:} The bottom portion presents a more challenging precision manipulation task. Notably, the strategy without post-processing fails to complete Stage-1 and Stage-2, as indicated by the ``Failed'' markers, highlighting that jerky motion not only slows down execution but can also cause task failures. Naive Switching successfully completes all stages with times of 4.93s, 3.87s, and 6.65s for Stage-1, Stage-2, and Stage-3 respectively, totaling 14.66s. VLA-RAIL achieves the fastest completion with 5.87s for Stage-3 and a total time significantly lower than both baseline methods. The failures observed in the without post-processing strategy underscore that trajectory smoothness is not merely a comfort metric but a critical factor for task success in precision manipulation.

\begin{table}[t]
\caption{Success Rate Across VLA Architectures. $\Delta$ denotes the absolute improvement from applying VLA-RAIL.}
\label{table_model_success}
\centering
\begin{tabular}{lccc}
\hline
Model & w/o post-processing &  VLA-RAIL & $\Delta$ \\
\hline
GO1 & 0.20 & 0.30 & +0.10 \\
SmolVLA & 0.15 & 0.45 & +0.30 \\
$\pi_0$ & 0.30 & 0.95 & +0.65 \\
$\pi_{0.5}$ & 0.225 & 0.95 & +0.725 \\
GR00T & 0.50 & 0.95 & +0.45 \\
\hline
\end{tabular}
\end{table}

\begin{figure*}[htbp]
  \centering
  \includegraphics[width=1\linewidth]{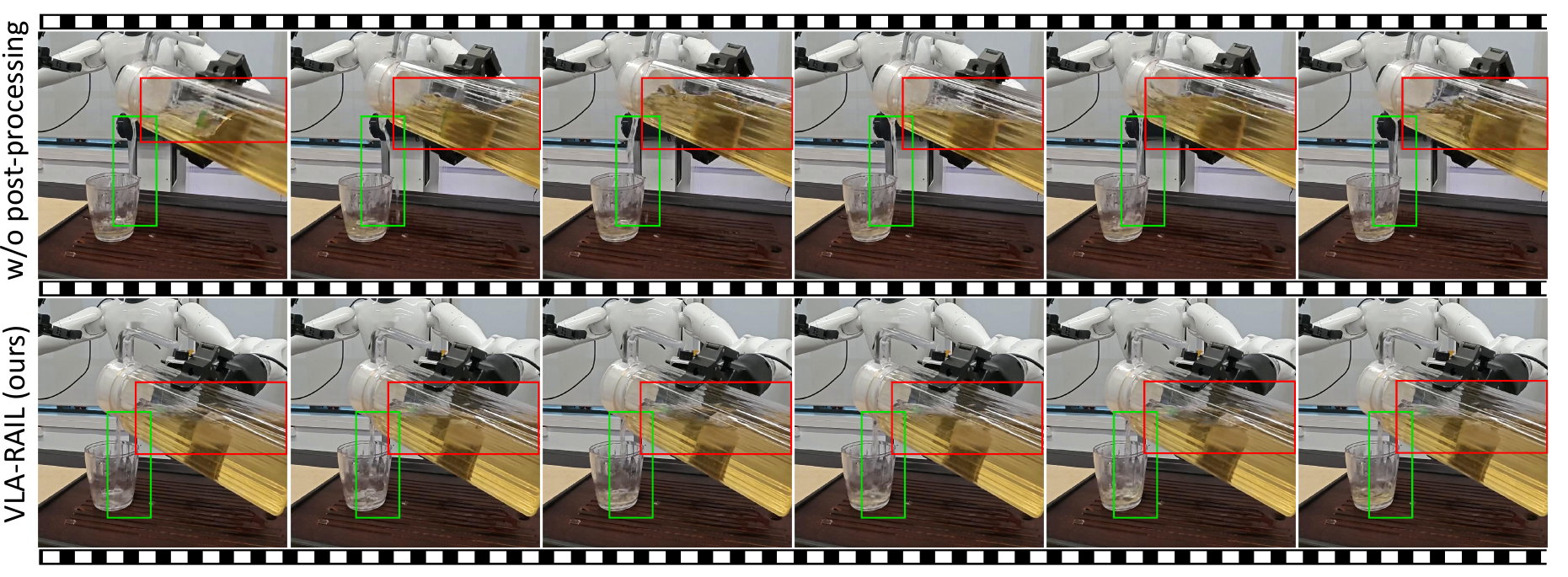}
  \caption{Qualitative comparison on the tea-pouring task. The top row displays robot executions driven by the raw VLA model outputs (w/o post-processing), whereas the bottom row displays executions with VLA-RAIL (ours). Motion jitter causes the liquid surface inside the teapot to oscillate and results in a discontinuous, splashing water stream during pouring. In contrast, smooth robot motion leads to more stable and continuous fluid motion.}
  \label{fig:qualitative}
\end{figure*}

\subsection{Compatibility across Multiple VLA Models}

VLA-RAIL is designed to be model-agnostic and therefore compatible with a wide range of VLA models. To validate this property, we evaluate five VLA models based on different generative paradigms on the same manipulation task, where the robot needs to pick up a bottle and place it into a basket. Table~\ref{table_model_success} summarizes the overall success rates without any post-processing and with VLA-RAIL.

The results show that VLA-RAIL consistently improves success rates across all evaluated models, with absolute gains ranging from +0.10 to +0.725. After applying VLA-RAIL, the models can be grouped into two distinct performance tiers: 1) High-performing models ($\pi_0$, $\pi_{0.5}$, and GR00T) all reach a success rate of 0.95 with VLA-RAIL, despite exhibiting substantially different baseline performance (0.30, 0.225, and 0.50, respectively). This indicates that these models generate fundamentally correct action sequences, whose execution quality is primarily degraded by trajectory-level noise, which VLA-RAIL effectively suppresses. 2) Lower-performing models (GO1 and SmolVLA) also benefit from VLA-RAIL, with SmolVLA improving from 0.15 to 0.45 and GO1 from 0.20 to 0.30. However, the more limited gains suggest that their failures are likely dominated by semantic-level errors rather than trajectory noise alone.

Notably, $\pi_{0.5}$ exhibits the largest improvement ($\Delta = +0.725$), suggesting that its generative process introduces significant stochastic noise that VLA-RAIL is particularly effective at filtering. These results confirm that VLA-RAIL serves as a general-purpose enhancement layer applicable across diverse VLA architectures without requiring model-specific tuning.

\subsection{Qualitative Results}
Fig.~\ref{fig:qualitative} presents a qualitative comparison of a fine-grained robotic manipulation task, showing consecutive frames from the tea-pouring task. When the robot is driven by the raw outputs of the VLA model, the liquid surface inside the teapot exhibits pronounced oscillations, and the water stream pouring into the cup suffers from severe discontinuities, splashing, and instability. In contrast, VLA-RAIL produces smooth and stable fluid motion, clearly demonstrating the advantages of our framework.

\section{Discussion and Conclusion}
In this work, we present VLA-RAIL, a robust framework to bridge high-latency VLA models and real-time robot control. By decoupling inference from execution and employing a novel quintic polynomial blending strategy, we achieved smooth, continuous motion that masks inference delays. Our experiments demonstrated significant improvements in task success rates and motion smoothness compared to standard synchronous and naive asynchronous baselines.

\textbf{Limitations:} One limitation of our current approach is the fixed parameterization of the transition duration. In highly dynamic scenarios, a fixed transition time may not be optimal. Precise alignment between observations and actions can only be achieved if VLA-RAIL and the robot hardware are tightly clock-synchronized. Moreover, although VLA-RAIL is already compatible with different robots, this advantage has not yet been validated due to cost constraints.

\textbf{Future Work:} Future research will focus on adaptive transition strategies that dynamically adjust the blending horizon based on the robot's velocity and the confidence of the VLA model. We also plan to integrate Model Predictive Control (MPC) at the client side to enforce safety constraints more explicitly during the transition phase.


\bibliographystyle{ieeetr}
\bibliography{scholar}
\end{document}